# A Novel Memetic Feature Selection Algorithm

Mohadeseh Montazeri

Department of Computer Engineering and Information Technology, Payame Noor University
Tehran, Iran
Department of Computer, Technical and Vocational University
Kerman, Iran
*mohadeseh_montazeri@yahoo.com*

Hamid Reza Naji

International Center for Science, High Technology and Environmental Sciences, Kerman Graduate University of Technology
Kerman, Iran
*hamidnaji@ieee.org*

Mitra Montazeri

Medical Informatics Research Center, Institute for Futures Studies in Health, Kerman University of Medical Sciences,
Kerman, Iran
*mmontazeri@eng.uk.ac.ir*

Ahmad Faraahi

Department of Computer Engineering and Information Technology, Payame Noor University
Tehran, Iran
*afaraahi@pnu.ac.ir*

*Abstract*— **Feature selection is a problem of finding efficient features among all features in which the final feature set can improve accuracy and reduce complexity. In feature selection algorithms search strategies are key aspects. Since feature selection is an NP-Hard problem; therefore heuristic algorithms have been studied to solve this problem.**

**In this paper, we have proposed a method based on memetic algorithm to find an efficient feature subset for a classification problem. It incorporates a filter method in the genetic algorithm to improve classification performance and accelerates the search in identifying core feature subsets. Particularly, the method adds or deletes a feature from a candidate feature subset based on the multivariate feature information. Empirical study on commonly data sets of the university of California, Irvine shows that the proposed method outperforms existing methods.**

I. INTRODUCTION

Most of real world classification problems require supervised [1] or semi-supervised [2] learning which use class information to establish a model to predict unseen instance. In these models, the underlying class probabilities and class conditional probabilities are unknown [3], and some instances are associated with a class label. In real-world situations, relevant features are often unknown a priori. Therefore, many candidate features are introduced to better represent the domain. In theory, more features should provide more discriminating power, but in practice, with a limited amount of training data, excessive features will not only significantly slow down the learning process, but also cause the classifier to over-fit the training data as uninformative features may confuse the learning algorithm. Feature Selection is the problem of selecting informative feature among all features in which a selected feature subset has lower cardinality and have higher accuracy. Feature selection has been an active and fruitful field of research and development for decades in statistical pattern recognition [4], machine learning [5, 6], data mining [7, 8] and statistics [9, 10].

In general, a feature selection algorithm consists of four basic steps: subset generation, subset evaluation, stopping criterion, and result validation [11]. Subset generation is a search procedure. Basically, it generates subsets of features for evaluation. Let $N$ denote the number of features in the original data set, then the total number of candidate subset is $2^N$ which makes exhaustive search through the feature space infeasible with even moderate $N$. Each subset generated by the generation procedure needs to be evaluated by a certain evaluation criterion and compared with the previous best one with respect to this criterion. If it is found to be better, then it replaces the previous best subset. Without a suitable stopping criterion the feature selection process may run exhaustively before it stops. The selected best feature subset needs to be validated on both the selected subset and the original set and comparing the results using artificial data sets and/or real-world data sets.

Researchers have studied various aspects of feature selection. There are two key aspects of feature selection: feature evaluation and search strategies. Feature evaluation is how to measure the goodness of a feature subset [12, 13]. There are filter models [14-16] and wrapper models [17-19] with different emphases on dimensionality reduction or accuracy enhancement.

The filters approach is based on the intrinsic properties of the data. The essence of filters is to seek the relevant features and eliminate the irrelevant ones. This method finds efficient features in one of two ways: univariate method and multivariate method. In univariate method, the idea here is to compute some simple score *S(i)* that measures how informative each feature $x_i$ is about the class labels *y*. using this method have three problems. First problem, features that are not individually relevant should become deleted but they may become relevant in the context of others [20]; second problem, always the relevant feature is not useful one because of possible redundancies [21]. Third problem, when the features were ranked according to their scores *S(i)*, if the number of effective feature is not determine, decide how many features is difficult and time consuming.  Therefore, the second method in filter approach is attended. In this method, it takes into account feature dependencies. This method potentially achieves better results because they consider feature dependence [21] but it is obvious, they need search strategies in feature space to find the best feature subset. All filter based methods are fast and allow them to be used with any learner. An important disadvantage of filter methods is that they ignore the interaction with the classifier therefore they have low accuracy.

Beside this method, wrapper approach is placed. Wrapper approach is a procedure that "wraps" around a learning algorithm, and accuracy of this learning algorithm is to be used to evaluate the goodness of different feature subsets. It is slow to execute because the learning algorithm is called repeatedly, but it has high accuracy. Another class is embedded methods which incorporate feature subset generation and evaluation in the training algorithm. This method like wrapper approach does not separate the learning from the feature selection part but incorporating knowledge about the specific structure of the classification or regression function. In actual, the structure of the class of functions under consideration plays a crucial role. Recently, hybrid models are proposed to combine the advantages of both filter models.

Filters and wrappers differ mostly by the evaluation criterion. Both filter and wrapper methods can make use of search strategies to explore the space of all possible feature combinations that is usually too large to be explored exhaustively. Therefore, another important aspect i.e. search strategies [12] were studied by researchers.

Heuristic search employs heuristics in conducting search. Due to polynomial complexity, it can be implemented very faster than previous searches. These methods do not guarantee the optimality or feasibility. Therefore many researches have been done to increase and guarantee the optimality.

A meta-heuristic algorithm, which is kind of heuristic algorithm,  by tightening a focus on good solutions and improving upon them (intensification), and to encourage the exploration of the solution space by broadening the focus of the search into new areas (diversification) can search solution space effectively [23]. Exploration and Exploitation are two competing goals govern the design of global search methods. Exploration is important to ensure global reliability, i.e., every part of the domain is searched enough to provide a reliable estimate of the global optimum; exploitation is also important since it concentrates the search effort around the best solutions found so far by searching their neighborhoods to produce better solutions [37]. Meta-heuristc is capable of global exploration and locating new regions of the solution space to identify potential candidates, but there is no further focus on the exploitation aspect when a potential region is identified [24]. Thus, Memetic Algorithms (MA) which incorporate local improvement search into meta-heuristics, were proposed [25]. Experimental studies have been shown that a hybrid of a meta-heuristic and a local search is capable of more efficient search capabilities [26].

In this paper, we propose a feature selection algorithm based on memetic algorithm (MFS). The goal of our method is to improve classification performance and accelerate the search to identify important feature subsets.  In particular, the filter method tunes the population of Genetic Algorithm (GA) solutions by adding or deleting features based on multivariate feature information. Hence, our focus is on filter methods that are able to assess the goodness of one feature in the context of others. We denote popular filter method, Pearson Correlation Coefficient, as our filter method in this correspondence. Furthermore, we investigate the balance between exploitation and exploration. Empirical study of our method on most commonly used data sets from the University of California, Irvine (UCI) repository [27] indicates that it outperforms recent existing methods. The rest of this article is organized as follow: Section 2 describes memetic algorithm. In Section 3, we explain MFS. The experimental results and conclusion are presented in Section 4 and 5, respectively.

## II.    MEMETIC ALGORITHM

Memetic algorithms [25] are population-based meta-heuristic  search approaches that have been receiving increasing attention in the recent years. They are inspired by Neo-Darwinian's principles of natural evolution and Dawkins' notion of a meme defined as a unit of cultural evolution that is capable of local refinements. Generally, Memetic algorithms  may be regarded as a marriage between a population-based global search and local improvement procedures. It has shown to be a successful and popular method for solving optimization problems in many contexts [39].

Fig.1 represents general framework of memetic algorithm. As you can see, two important searches have done on a problem, global and local search. With these two searches, exploration and exploitation have done properly and hence solution space is searched effectively. Nevertheless, Meta-heuristc algorithms generally suffer from excessively slow convergence to locate a precise enough solution because of their failure to exploit local information.

   Memetic algorithm often limits the practicality of Meta-heuristc algorithms on many large-scale real world problems where the computational time is a crucial consideration.

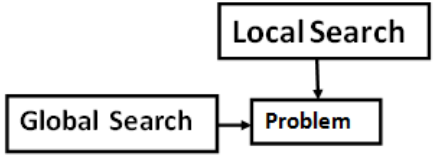

Figure 1. General framework of memetic algorithm

## III. MFS

Figure. 2 show the procedure of MFS. In the first step, the GA population is initialized randomly which each chromosome is encoding a candidate feature subset. Then, on the elite chromosomes, a local search or meme is applied. The mechanism of local improvement can be reaching a local optimum or improving the solution. Genetic operators are then used to generate the next population. This process repeats until the stopping conditions are satisfied.

The local search performs a search on complete solutions to improve their solutions. It causes improving the quality of the candidate solution at hand. The goal of local search is to produce a better candidate solution at each step to increase exploitation.

### A. Chromosome Encoding

In the MFS, encoding solution is a binary string with a length equal to the total number of features.
Each bit encodes a single feature. A bit of "1" implies the corresponding feature is selected and "0" is not so that the length of the chromosome is $N$.

### B. Fitness Function

To evaluate the goodness of each feature subset which is generated by each chromosome, we use accuracy of classification. It can be defined based on (1):

$$Fit = J(fs) \qquad (1)$$

Where *fs* denotes the corresponding selected feature subset encoded in solution, *J* compute the goodness of feature subset.

In this paper, it is accuracy of classification. Note that if two solutions have the same fitness, which one has smaller feature number is selected.

### C. Evolutionary Operations

Evolutionary operations are selection, crossover and mutation. In selection, we use a rank based elitism roulette wheel selection which is based on the fitness of chromosomes. It should ensure that fitter chromosomes have better chance to survive. We use one-point crossover such that if two parent chromosomes $C_1$ and $C_2$ are selected, they perform crossover operation with a crossover probability $P_c$ to generate two new chromosomes $Off_1$ and $Off_2$ with exchanging information in a randomly cut point. In mutation operator selects some positions with a mutation probability $P_m$ randomly and invert genes at these positions.

### D. The Local Search Improvement Procedure

Our local search iterate over each feature of candidate feature subset. At each step, one feature is deleted or added (it depend to candidate feature subset). New feature subset is evaluated, if an improvement is achieved, new feature subset is accepted as the current one. Then the iteration continues with another feature. This process iterates for *L* (LS steps) times.

Figure. 3 show the procedure of our local search procedure. In fact, both delete and add features are possible.

### E. Feature Subset Evaluation filter(FE)

In local search there is a filter evaluation function to evaluate solution which is generated during the local search procedure. In our method, for improving our work effectively, we used Pearson's correlation coefficient (3) which is a famous subset evaluation filter.

Correlation coefficients are the simplest approach to feature relevance measurements. The linear correlation coefficient of Pearson is very popular in statistics and represents the normalized measure of the strength of linear relationship between variables [28]. For random variable X with value x and random variable Y with value y it is defined as (2).

**Procedure of MFS**
1  **Begin**
2    **Initialize:** Randomly initialize population of feature subset, initialize E and others;
3    **While** (stop if condition is not satisfied)
4      Evaluate fitness of all feature subset encoded in the population;
5      Find E best feature subset in the population and put them into elite pop;
6      **For** (each subset in elite pop )
7        Perform local search and replace it with new feature subset;
8      **End For**
9      Evaluate fitness of new solutions which is generated by local search;
10     Select the best solution based on fitness function;
11     Perform evolutionary operators, i.e. selection, crossover, mutation;
12   **End While**
13 **End**

Figure 2. The procedure of MFS.

**Procedure of Local Search**
1  **Begin**
2    **Input:** Elite population;
3    **Initialize:** K;
4    **For** (each feature subset in elite population(E), $E_i$)
5      **For** (number of K)
6        $E_{best} = E_i$;
7        Add or delete each feature in $E_i$;
8        calculate filter evaluation of improved feature subset fs using FE(fs);

```
9         If (FE(fs)>FE(E_best))
10            E_best = fs;
11        Else
12            change feature subset in original format;
13        End If
14        Replace E_i with E_best;
15    End For
16  End For
17 End
```

Figure 3. The procedure of our local search method.

$$r_{x,y} = \frac{\sum_i (x_i - \bar{x}_i)(y_i - \bar{y}_i)}{\sqrt{\sum_i (x_i - \bar{x}_i)^2 \sum_j (y_i - \bar{y}_i)^2}} \quad (2)$$

Which is equal to ± 1 if X and Y are linearly dependent and zero if they are completely uncorrelated. Some random variables may be correlated positively, and some negatively.

If a group of k features variables has already been existed, correlation coefficients may be used to estimate correlation between this group and the class variable, including inter-correlations between the features. Relevance of a group of features grows with the correlation between features and classes, and decreases with growing inter-correlation. These ideas have been discussed in theory of psychological measurements [29] and in the literature on decision making and aggregating opinions [30]. In 1964, Ghiselli [29] proposed (3):

$$J_k = \frac{k\bar{r}_{cf}}{\sqrt{k + k(k-1)\bar{r}_{ff}}} \quad (3)$$

Where $\bar{r}_{cf}$ is the average correlation coefficient between these $k$ features and the output variables and the average between different features as $\bar{r}_{ff}$. This formula is obtained from Pearson's correlation coefficient with all variables standardized. It has been used in the Correlation-based Feature Selection (CFS) algorithm [31].

## IV. EXPERIMENTAL RESULT AND DISCUSSION

In this section, our experimental result is carried out to show the effectiveness of our method. In The following subsections, a brief description of dataset benchmark is given, and then our simulations results and comparison with literature works are presented and discussed.

### A. Database Description and preprocessing

We use 6 benchmark datasets which are frequently used in literatures. They are from the University of California, Irvine (UCI) repository [27]. Table I shows description of these datasets. They are both nominal and numerical data.

Since some of these datasets have missing values or continues values in uncontrolled rang, they have a preprocessing step before they are used. For missing values, we replaced them with the most frequently used values for nominal and numeric features. To control the range of continues features we normalize them in rang [0, 1].

- K-Fold Cross Validation: This metric is used to estimate how accurately a predictive model will perform in practice and represents the probability that an instance was classified correctly. In *k*-fold cross-validation, the original sample is randomly partitioned into *k* subsamples. A single subsample is retained as the test data and the remaining *k* −1 subsamples are used as training data. The cross-validation process is then repeated *k* times, with each of the *k* subsamples used exactly once as the validation data.

TABLE I.
TABLE II. DESCRIPTION OF DATASETS.

| No. | Database | N | Number of instances | Number of classes |
|---|---|---|---|---|
| 1. | Lymphography | 18 | 148 | 4 |
| 2. | isolet | 617 | 1559 | 26 |
| 3. | Synthetic | 60 | 600 | 6 |
| 4. | Audiology | 69 | 226 | 24 |
| 5. | Dermatology | 34 | 366 | 6 |
| 6. | Musk clean1 | 166 | 476 | 2 |

Then *k* results from the folds then can be averaged to produce a single estimation. In our study, due to being randomness, run 10 times and at each time a 10-fold cross validation which is commonly used is used [35], and the final results were their average values (10-10 fold CV).

### B. Performance evaluation

In this section, we present an experimental study of MFS on commonly used UCI data sets. We employed a population size of 30 and generation number is 200. Crossover rate, Pc, and mutation rate. Pm, are 0.6 and 0.1, respectively. Table II shows the best and average accuracy of 5 runs of MFS on defined databases. Because MFS is a random search algorithm, different results may be obtained at every run. We have run this algorithm on 5 runs and record average of them.

### C. Comparison Of literature Works

We empirically evaluated the performance of MFS by comparing with recently methods, Ref. [40] and Ref. [41]. We have compared our method with two typical feature selectors: ReliefF and IG. ReliefF [42] is a popular instance-based feature weighting algorithms and Information Gain (IG) measures the decrease in entropy when the features are presented [43]. They are all well-known methods and have excellent performance. For ReliefF, we use 5 neighbors and 30 instances throughout the experiments as suggested by Robnik-Sikonja and Kononenko [44], which is also used in the literature [45]. The results of comparisons are reported in table III. In each row best results are bolded. As we can see in this table, in most cases the presented method (MFS) has better results which are considerable in some databases.

TABLE III. PERFORMANCE OF MFS (1NN, 10-10 FOLD CV, UNIT: %)

| No. | Database | Accuracy of Unselected features | Best accuracy | Average accuracies |
|---|---|---|---|---|
| 1 | Lymphography | 83.78 | 87.23 | 85.34 |
| 2 | Isolet | 85.12 | 89.22 | 71.80 |
| 3 | Synthetic | 98.16 | 99.22 | 99.06 |
| 4 | Audiology | 74.04 | 80.53 | 79.89 |
| 5 | Dermatology | 95.36 | 97.21 | 97.05 |
| 6 | Musk clean1 | 89.28 | 92.65 | 91.66 |

TABLE IV. THE COMPARISON OF MFS WITH PREVIOUS WORKS (UNIT: %)

| NO. | Database | MFS | | Results obtained from other methods | | ReliefF | IG |
|---|---|---|---|---|---|---|---|
| 1 | Lymphography | MFS+1NN 10-10 fold CV | **87.23 (85.34)** | CoFS+1NN 10-fold CV Ref.[40] | 79.95 | 78.00 | 82.23 |
| 2 | Isolet | MFS+1NN 10-10 fold CV | **89.22 (71.80)** | CoFS+1NN 10-fold CV Ref.[40] | 66.52 | 54.65 | 53.62 |
| 3 | Synthetic | MFS+1NN 10-10 fold CV | **99.22 (99.06)** | CoFS+1NN 10-fold CV Ref.[40] | 92.00 | 86.17 | 83.00 |
| 4 | Audiology | MFS+1NN 10-10 fold CV | **80.53 (79.89)** | DMIFS+1NN 10-10 fold CV Ref. [41] | 74.04 | 72.94 | 72.32 |
| 5 | Dermatology | MFS+1NN 10-10 fold CV | **97.21 (97.05)** | DMIFS+1NN 10-10 fold CV Ref. [41] | 92.18 | 83.96 | 81.37 |
| 6 | Musk clean1 | MFS+1NN 10-10 fold CV | **92.65 (91.66)** | DMIFS+1NN 10-10 fold CV Ref. [41] | 87.34 | 84.10 | 83.53 |

## V. CONCLUSION

In this paper, we have proposed a novel method based on memetic algorithm to find an efficient feature subset. The goal of MFS was to improve classification performance and accelerate the search methodology to identify important feature subsets. MFS was based on a heuristic approach which can search the solution space effectively by appropriate exploring and exploiting. Because exploration and exploration are important properties in heuristic algorithms. MFS could dose a trade of between exploitation and exploration effectively. Exploration was done by genetic operators and exploitation by our local search. Our local search heuristics improve the quality of the candidate solution to produce a better candidate solution at each step. We used filter method as local search heuristic. In addition, intensive local search can trap algorithm into local optimum but MFS controlled this issue by study on number of iteration in local search heuristic (parameter $k$ and $E$).

MFS which was a wrapper-filter method was compared to well known methods. Empirical study of MFS on commonly used data sets from UCI data sets indicated that it outperformed recent methods.